\documentclass[11pt,a4paper]{article}
\usepackage{times,latexsym}
\usepackage{url}
\usepackage[T1]{fontenc}

\usepackage{microtype}

\usepackage{amsmath,amssymb,amsfonts,bm}
\usepackage{booktabs,multirow}
\usepackage{graphicx}
\usepackage{caption}
\usepackage{subcaption}

\usepackage{algorithm}
\usepackage{algorithmic}
\usepackage{float}
\usepackage{enumitem}
\usepackage{tcolorbox}
\usepackage{verbatim}
\usepackage{tcolorbox}
\tcbuselibrary{most}
\usepackage{listings}
\usepackage{xcolor}
\usepackage{xurl}   
\usepackage{hyperref}

\lstdefinestyle{promptstyle}{
  basicstyle=\ttfamily\footnotesize,
  columns=fullflexible,
  breaklines=true,
  breakatwhitespace=true,
  keepspaces=true,
  showstringspaces=false,
  upquote=true
}
\usepackage[acceptedWithA]{tacl2021v1}

\usepackage{xspace,mfirstuc,tabulary}

\newif\iftaclinstructions
\taclinstructionsfalse 
\iftaclinstructions
\renewcommand{\confidential}{}
\renewcommand{\anonsubtext}{(No author info supplied here, for consistency with
TACL-submission anonymization requirements)}
\newcommand{\instr}
\fi

\iftaclpubformat 

\else

\fi

\newcommand{\tss}{\ensuremath{\mathrm{TSS}}}
\newcommand{\tsspp}{\ensuremath{\mathrm{TSS{+}{+}}}}
\newcommand{\pmodel}{\ensuremath{p_\theta}}

\title{Task--Specificity Score: Measuring How Much Instructions Really Matter for Supervision}

\author{
\textbf{Pritam Kadasi\textsuperscript{1}\Thanks{Corresponding author.}} \quad
Abhishek Upperwal\textsuperscript{2} \quad
Mayank Singh\textsuperscript{1} \\
\textsuperscript{1}Lingo Research Group, Indian Institute of Technology Gandhinagar, India \\
\textsuperscript{2}Soket AI, India \\
\small{\texttt{\{pritam.k, singh.mayank\}@iitgn.ac.in}} \\
\small{\texttt{abhishek@soket.ai}}
}

\date{}

\begin{document}
\maketitle

\begin{abstract}
Instruction tuning is now the default way to train and adapt large language models, but many instruction--input--output pairs are only weakly specified: for a given input, the same output can remain plausible under several alternative instructions. This raises a simple question: \emph{does the instruction uniquely determine the target output?}

We propose the \textbf{Task--Specificity Score (TSS)} to quantify how much an instruction matters for predicting its output, by contrasting the true instruction against plausible alternatives for the same input. We further introduce \textbf{TSS++}, which uses hard alternatives and a small quality term to mitigate easy-negative effects. Across three instruction datasets (\textsc{Alpaca}, \textsc{Dolly-15k}, \textsc{NI-20}) and three open LLMs (Gemma, Llama, Qwen), we show that selecting task-specific examples improves downstream performance under tight token budgets and complements quality-based filters such as perplexity and IFD.
\end{abstract}

\section{Introduction}
\label{sec:intro}

Instruction tuning has become a central paradigm for adapting pretrained language models to follow natural-language instructions across a wide range of tasks \citep{ouyang2022training,wang-etal-2023-self-instruct,taori2023alpaca}. By fine-tuning on large mixtures of instruction--input--output examples, models acquire broad task generality and improved alignment with user intent. As instruction datasets continue to grow in size and diversity, however, a fundamental question remains underexplored: \emph{which instruction examples actually provide task-defining supervision, and which are largely redundant?}

Most existing approaches to instruction data selection and weighting focus on \emph{output quality} or \emph{model fit}. Perplexity-based filtering prefers examples with high likelihood under a pretrained model, implicitly favoring fluent, in-distribution responses \citep{zhou2023lima, li-etal-2024-superfiltering}. Instruction-following difficulty (IFD) variants compare likelihood with and without the instruction to estimate whether the instruction helps predict the output \citep{li-etal-2024-quantity}. More recent work studies data selection through gradient similarity, influence functions, or representational diversity \citep{pruthi2020learning,xia2024less, zhang2025harnessing, yeh2022first}. While effective in many settings, these signals answer a different question than the one practitioners often care about: \emph{does this example actually require its instruction to specify the task being solved?}

In practice, many instruction datasets contain weakly specified supervision. For a given input $X$, an output $Y$ may be compatible with multiple plausible instructions: a factual answer might satisfy both ``answer the question'' and ``summarize the passage,'' or a short label might fit several classification prompts. Such examples are easy for models to learn but provide limited information about how instructions constrain behavior. Training heavily on these instances can waste budget and obscure task boundaries, especially in budget-constrained settings or when mixing heterogeneous instruction sources.

In this paper, we propose the \emph{Task--Specificity Score} ($\tss$), a simple and model-agnostic measure of how uniquely an output $Y$ is specified by its instruction $I$ among other plausible instructions for the same input $X$. Rather than asking whether an example is high-quality or easy for the model, $\tss$ asks a different question: \emph{would this output still make sense if the instruction were changed?} Intuitively, an example has high task specificity if $Y$ is strongly compatible with the given instruction and significantly less compatible with reasonable alternative instructions that could have been asked for the same input.

We operationalize this idea by contrasting the log-likelihood of the output under the true instruction with its likelihood under a set of counterfactual instructions generated without access to the output. This yields a pointwise estimate of how informative the instruction--output pairing is, relative to nearby tasks. From an information-theoretic perspective, $\tss$ can be viewed as a Monte Carlo approximation to the conditional mutual information between instruction and output given the input, $I(I;Y\mid X)$, which is otherwise intractable for large language models.

Building on this foundation, we introduce \textbf{$\tsspp$}, a quality-aware contrastive extension designed to be robust on structured instruction corpora. $\tsspp$ combines (i) a normalized, InfoNCE-style contrastive objective computed against \emph{hard} alternative instructions and (ii) an explicit likelihood-based quality term that captures in-distribution fluency. This formulation addresses two practical failure modes of vanilla specificity measures: sensitivity to easy negatives and underweighting of output quality, which has made simple perplexity filtering surprisingly strong in prior work \citep{zhou2023lima,raffel2020exploring}.

We evaluate $\tss$ and $\tsspp$ as filtering and weighting mechanisms for instruction tuning under explicit token budgets. Across three widely used instruction datasets (Alpaca, Dolly-15k, and a 20-task subset of Natural Instructions) and three open-weight decoder-only models (Gemma, LLaMA, and Qwen), we measure downstream performance on four diverse benchmarks: ARC-C, MMLU, HellaSwag, and TruthfulQA. Our experiments show that selecting or upweighting high-specificity examples consistently improves budget efficiency and often outperforms strong quality-based baselines, particularly at low retention fractions. Notably, $\tsspp$ remains competitive with perplexity and IFD filters on structured datasets while providing a more principled notion of task informativeness.

Our contributions are threefold:
\begin{itemize}[nolistsep]
    \item We introduce \textbf{Task--Specificity Score ($\tss$)}, a principled, instance-level measure of how much an instruction actually constrains the output relative to plausible alternative tasks.
    \item We propose \textbf{$\tsspp$}, a contrastive, quality-aware extension that combines hard-negative instruction comparisons with likelihood-based quality control.
    \item Through extensive experiments across models, datasets, and budget regimes, we demonstrate that task specificity is a complementary and practically useful axis for instruction data selection, improving downstream performance and data efficiency beyond standard quality-based heuristics.
\end{itemize}

Taken together, our results suggest that not all instruction examples contribute equally to learning how instructions matter. Measuring and exploiting task specificity provides a new lens on supervision quality, one that is especially relevant as instruction tuning continues to scale under finite compute and data budgets.

\section{Related Work}
\label{sec:related}

\paragraph{Instruction tuning and mixture construction.}
Instruction tuning has become the dominant paradigm for aligning pretrained LMs to follow natural-language directives, popularized by RLHF-style instruction-following models \citep{ouyang2022training} and large-scale supervised instruction mixtures \citep{wei2022finetuned,wang-etal-2023-self-instruct}. Broad instruction corpora such as \textsc{Natural Instructions} \citep{mishra-etal-2022-cross} and subsequent community datasets (e.g., Alpaca \citep{taori2023alpaca}) motivated training on heterogeneous datasets where the same input can admit many plausible tasks. In practice, these datasets are often mixed using simple heuristics (uniform or size-proportional sampling), which can be effective but do not explicitly measure whether the instruction is actually necessary to recover the labeled output for a given example.

\paragraph{Data filtering/selection for instruction tuning.}
A long line of work studies dataset selection under limited budgets, including classical cross-entropy difference and domain-adaptation selection \citep{moore-lewis-2010-intelligent,axelrod-etal-2011-domain}. Recent instruction-tuning settings revisit selection with model-based signals and learned filters: \citet{li-etal-2024-quantity} propose self-guided data selection to trade quantity for quality during instruction tuning, while \citet{li-etal-2024-superfiltering} study learned ``superfilters'' that can remove low-utility instruction data at scale. Orthogonally, \citet{xia2024less} frame \emph{targeted} instruction tuning as selecting influential examples for a desired capability via optimizer-aware influence approximations. A complementary survey and taxonomy of instruction-tuning data selection and curation is provided by \citet{liu-etal-2025-take}. Our work fits this literature but focuses on a different axis: isolating \emph{instruction utility} from generic quality, i.e., whether the instruction meaningfully constrains the supervision beyond what other plausible instructions for the same input would support.


\paragraph{Measuring instruction utility beyond quality.}
Many practical filters for instruction tuning primarily track output ``quality'' \citep{zhou2023lima, chen2024alpagasus} or in-distributionness (e.g., likelihood/perplexity under a reference model) \citep{li-etal-2024-superfiltering}, or compare likelihood with vs.\ without the instruction (instruction-following difficulty style signals) \citep{li-etal-2024-quantity}. These methods are valuable but can conflate (i) high-quality outputs that are compatible with many tasks and (ii) examples where the instruction uniquely determines the label. Our \emph{Task--Specificity Score} instead operationalizes instruction utility through \emph{counterfactual instructions}: we compare the likelihood of the observed output under the true instruction against its likelihood under a set of plausible alternatives for the same input, connecting to contrastive learning views of discrimination among negatives \citep{oord2019representationlearningcontrastivepredictive}. This explicitly targets when instructions ``really matter'' for supervision, and is therefore complementary to quality-centric selection \citep{zhou2023lima} and learned filtering methods \citep{xia2024less, tu-etal-2025-resofilter, xu-etal-2025-stronger}.

\section{Method: Task--Specificity Score ($\tss$)}
\label{sec:method}

\paragraph{Setup.}
We assume a corpus of instruction--input--output triples
$\mathcal{D}=\{(I,X,Y)\}$, where $I$ is a natural-language instruction,
$X$ is the input (e.g., question/passage), and $Y$ is the target output (e.g., answer/label/summary).
Let $\pmodel$ be a pretrained (frozen) language model that assigns conditional probabilities
$p_\theta(Y\mid I,X)$ under the standard LM factorization over tokens.

\paragraph{Motivation.}
Many data selection signals used in instruction tuning are \emph{quality} proxies.
For example, perplexity-based filtering prefers examples with high likelihood under the model,
and instruction-following difficulty (IFD)~\cite{li-etal-2024-quantity} variants compare likelihood \emph{with} vs.\ \emph{without} the instruction.
These answer questions like: \emph{``Is $Y$ probable under $(I,X)$?''} or \emph{``Does adding $I$ help predict $Y$?''}
In contrast, we aim to measure \emph{task specificity}: whether the pair $(I,Y)$ is uniquely informative
among other plausible instructions that could reasonably be asked for the same input $X$.

\begin{algorithm}[t]
\caption{Computing Task--Specificity Score (TSS)}
\label{alg:tss}
\small
\begin{algorithmic}[1]
\REQUIRE Dataset $\mathcal{D}=\{(I_j,X_j,Y_j)\}_{j=1}^N$, frozen model $\pmodel$, \#alternatives $K$, retention fraction $\rho$
\FOR{each example $(I,X,Y)\in\mathcal{D}$}
  \STATE Sample $K$ alternative instructions $\{I_1,\dots,I_K\}$ using $\texttt{Prompt}(X)$ (no access to $Y$)
  \STATE $s_{\text{true}} \gets \log p_\theta(Y\mid I,X)$
  \FOR{$k=1$ to $K$}
    \STATE $s_k \gets \log p_\theta(Y\mid I_k,X)$
  \ENDFOR
  \STATE $\tss(I,X,Y) \gets s_{\text{true}} - \log\!\left(\frac{1}{K}\sum_{k=1}^K e^{s_k}\right)$
\ENDFOR
\STATE Threshold $\tau_\rho \leftarrow (1-\rho)$-quantile of $\{\tss\}$
\STATE Retained set $\mathcal{D}_\rho \leftarrow \{(I,X,Y): \tss(I,X,Y)\ge\tau_\rho\}$
\RETURN $\mathcal{D}_\rho$ (or weights $w=\exp(\lambda\cdot\tss)$)
\end{algorithmic}
\end{algorithm}

\subsection{Generating plausible alternative instructions}
\label{sec:alts}
For each triple $(I,X,Y)$, we construct a set of $K$ \emph{plausible alternative instructions}
$\{I_1,\dots,I_K\}$ that could be asked about the same input $X$ \emph{without observing $Y$}.
We use a prompt template\footnote{Refer~\ref{sec:prompts} for exact prompt template.} that depends only on $X$, e.g.,
\emph{``Generate $K$ different instructions a user might give about this input, each asking for a different kind of output.''}
Denoting this template as $\texttt{Prompt}(X)$, we sample
$I_k \sim p_\theta(\cdot \mid \texttt{Prompt}(X))$ for $k=1,\dots,K$.

\paragraph{Rationale.}
These alternatives form a local, model-induced approximation to the distribution over
``other tasks one might ask for the same input''. Crucially, they are constructed without conditioning on $Y$.

\subsection{Task--Specificity Score ($\tss$)}
\label{sec:tss}
Given the true triple $(I,X,Y)$ and its alternatives $\{I_k\}_{k=1}^K$,
we compare the compatibility of $Y$ with the true instruction against its compatibility under counterfactual instructions.
Let $s_{\text{true}}=\log p_\theta(Y\mid I,X)$ and $s_k=\log p_\theta(Y\mid I_k,X)$.
We define the Task--Specificity  (see Algorithm~\ref{alg:tss}):

\begin{equation}
\scalebox{0.60}{$
\begin{aligned}
\tss(I,X,Y)
= s_{\text{true}}-\log\!\left(\frac{1}{K}\sum_{k=1}^{K} e^{s_k}\right)
= \log \frac{p_\theta(Y\mid I,X)}{\frac{1}{K}\sum_{k=1}^K p_\theta(Y\mid I_k,X)}.
\end{aligned}
$}
\label{eq:tss}
\end{equation}

\paragraph{Interpretation.}
A high $\tss$ indicates that $Y$ is substantially more likely under the true instruction than under other plausible instructions for the same input.
A low (or negative) $\tss$ suggests that $Y$ remains similarly plausible under many alternative tasks, implying weakly specified supervision.

\subsection{Information-Theoretic view}
\label{sec:mi}
Conceptually, we care about how much information the instruction provides about the output beyond the input alone,
captured by conditional mutual information $I(I;Y\mid X)$.
Pointwise, this resembles $\log \tfrac{p(Y\mid I,X)}{p(Y\mid X)}$, but $p(Y\mid X)$ is intractable for LMs.
$\tss$ can be viewed as a Monte Carlo approximation where $p_\theta(Y\mid X)$ is estimated by averaging over alternative instructions:
$p_\theta(Y\mid X)\approx \tfrac{1}{K}\sum_{k=1}^K p_\theta(Y\mid I_k,X)$, which yields Eq.~\ref{eq:tss}.

\subsection{Using $\tss$ for Budgeted Filtering and Weighting}
\label{sec:use}
Once $\tss(I,X,Y)$ is computed for each training instance, we use it in two ways.

\paragraph{Filtering under a token budget.}
Given a retention fraction $\rho\in(0,1]$, we keep the top-$\rho$ fraction of examples by $\tss$.
Equivalently, let $\tau_\rho$ be the $(1-\rho)$-quantile of the $\tss$ scores; retain
$\mathcal{D}_\rho=\{(I,X,Y)\in\mathcal{D}:\tss(I,X,Y)\ge \tau_\rho\}$.
This implements budgeted selection (e.g., $\rho=0.05$ for 5\% of data/tokens).

\paragraph{Importance weighting.}
Alternatively, we define a nonnegative weight per example as $w=\exp(\lambda\cdot \tss)$ with scale $\lambda\ge 0$,
and train with a weighted loss $\mathbb{E}[w\cdot \ell_\theta(I,X,Y)]$.
Exponentiation provides a smooth, monotone mapping from score to positive weights and emphasizes high-TSS examples;
in practice, $\lambda$ is tuned and weights may be clipped for stability.

\subsection{$\tsspp$: Quality-aware contrastive specificity}
\label{sec:tsspp}
On structured instruction corpora (e.g., NI/Dolly-style), two practical issues can reduce the effectiveness of vanilla $\tss$:
(i) \emph{easy} or stylistically mismatched alternatives can make the denominator too small, inflating $\tss$,
and (ii) specificity alone can under-emphasize output \emph{quality} (fluency / in-distributionness), where strong PPL filters often excel.
We address this with \tsspp, which combines (a) a contrastive (InfoNCE-style)~\cite{oord2019representationlearningcontrastivepredictive} specificity term computed against \emph{hard} alternatives,
and (b) an explicit quality term.

\begin{algorithm}[t]
\caption{TSS++: Quality-Aware Contrastive Task Specificity}
\label{alg:tsspp}
\small
\begin{algorithmic}[1]
\REQUIRE Example $(I,X,Y)$, frozen model $\pmodel$, candidate dataset $\mathcal{C}(X)$, hard set size $m$, temperature $\tau$, quality weight $\alpha$
\STATE Compute normalized scores $s(I',X,Y)=\frac{1}{|Y|}\log p_\theta(Y\mid I',X)$ for all $I'\in\mathcal{C}(X)$
\STATE Select hard alternatives $\mathcal{H}(X)\leftarrow \textsc{Top-}m$ instructions in $\mathcal{C}(X)\setminus\{I\}$ by $s(I',X,Y)$
\STATE $s^+ \gets s(I,X,Y)$
\STATE $\displaystyle
\tss_{\mathrm{NCE}} \gets
\log\frac{\exp(s^+/\tau)}
{\exp(s^+/\tau)+\sum_{I^-\in\mathcal{H}(X)}\exp(s(I^-,X,Y)/\tau)}
$
\STATE $\tsspp(I,X,Y) \gets \tss_{\mathrm{NCE}} + \alpha\cdot s^+$
\RETURN $\tsspp(I,X,Y)$ (for filtering or weighting)
\end{algorithmic}
\end{algorithm}

\paragraph{Hard alternatives.}
Instead of using only random samples, we construct a candidate dataset $\mathcal{C}(X)$ of alternative instructions
and keep the most confusable ones.
In practice, $\mathcal{C}(X)$ can be generated alternatives, retrieved in-distribution instructions using KNN over embeddings of original instruction,
or a  of both; we then select the hard set $\mathcal{H}(X)\subset \mathcal{C}(X)$ as those with highest
average log-likelihood of $Y$ under the alternative instruction.

\paragraph{Normalized log-likelihood.}
We use length-normalized log-likelihood
$s(I,X,Y)=\frac{1}{|Y|}\log p_\theta(Y\mid I,X)$ to reduce sensitivity to output length, especially when $Y$ is short.

\paragraph{Contrastive specificity (InfoNCE form).}
Given temperature $\tau>0$ and hard alternatives $\mathcal{H}(X)$, define

\begin{equation}
\scalebox{0.60}{$
\begin{aligned}
\tss_{\mathrm{NCE}}(I,X,Y)
&= \log \frac{\exp\!\big(s(I,X,Y)/\tau\big)}
{\exp\!\big(s(I,X,Y)/\tau\big)
+\sum_{I^{-}\in \mathcal{H}(X)} \exp\!\big(s(I^{-},X,Y)/\tau\big)} \\
&= -\log\!\Bigg(
1+\sum_{I^{-}\in \mathcal{H}(X)}
\exp\!\Big(\big(s(I^{-},X,Y)-s(I,X,Y)\big)/\tau\Big)
\Bigg).
\end{aligned}
$}
\label{eq:tssnce}
\end{equation}

This normalization makes the score depend on \emph{relative} compatibility between the true instruction and hard alternatives.
Hard negatives are essential: if alternatives are too easy (very low likelihood), the contrast provides little discrimination.

\paragraph{Quality-aware $\tsspp$.}
Finally, we add an explicit quality term (the in-distribution likelihood under the true instruction) (see Algorithm~\ref{alg:tsspp}):
\begin{equation}
\scalebox{0.80}{$
\begin{aligned}
\tsspp(I,X,Y) \;=\; \tss_{\mathrm{NCE}}(I,X,Y) \;+\; \alpha \cdot s(I,X,Y),
\end{aligned}
$}
\label{eq:tsspp}
\end{equation}
where $\alpha\ge 0$ trades off instruction-specificity (contrastive) against output quality (likelihood).
We use $\tsspp$ in the same way as $\tss$: filter top-$\rho$ examples or compute weights $w=\exp(\lambda\cdot \tsspp)$.


\section{Experimental Setup}
\label{sec:exp_setup}

\subsection{Data}

\begin{table}[tb]
\centering
\tiny
\begin{tabular}{lrrr}
\toprule
Category & \#Tasks & \#Instances & \#Tokens \\
\midrule
Translation                 & $394$ & $1{,}182{,}213$ & $72{,}549{,}385$ \\
Question Answering          & $207$ & $  470{,}108$   & $106{,}180{,}992$ \\
Program Execution           & $ 90$ & $  433{,}157$   & $35{,}066{,}354$  \\
Sentiment Analysis          & $ 75$ & $  253{,}432$   & $32{,}670{,}340$  \\
Question Generation         & $ 83$ & $  230{,}103$   & $56{,}131{,}362$  \\
Text Matching               & $ 43$ & $  173{,}171$   & $14{,}178{,}609$  \\
Text Categorization         & $ 46$ & $  154{,}556$   & $11{,}876{,}178$  \\
Commonsense Classification  & $ 24$ & $  130{,}524$   & $ 2{,}237{,}835$  \\
Toxic Language Detection    & $ 40$ & $  115{,}584$   & $ 5{,}148{,}102$  \\
Fill in The Blank           & $ 22$ & $   93{,}210$   & $13{,}687{,}063$  \\
Textual Entailment          & $ 27$ & $   92{,}651$   & $ 4{,}886{,}841$  \\
Information Extraction      & $ 34$ & $   91{,}850$   & $ 6{,}710{,}148$  \\
Text Completion             & $ 21$ & $   86{,}145$   & $ 8{,}631{,}760$  \\
Sentence Perturbation       & $ 15$ & $   80{,}789$   & $ 2{,}472{,}566$  \\
Title Generation            & $ 19$ & $   80{,}696$   & $20{,}869{,}481$  \\
Wrong Candidate Generation  & $ 27$ & $   73{,}546$   & $ 9{,}763{,}136$  \\
Sentence Composition        & $ 20$ & $   72{,}496$   & $ 5{,}066{,}324$  \\
Question Understanding      & $ 16$ & $   63{,}448$   & $ 3{,}239{,}390$  \\
Pos Tagging                 & $ 10$ & $   62{,}118$   & $ 2{,}583{,}225$  \\
Summarization               & $ 16$ & $   59{,}200$   & $43{,}445{,}932$  \\
\bottomrule
\end{tabular}
\caption{Natural Instructions dataset statistics by task category.}
\label{tab:dataset}
\end{table}

We study three instruction-following corpora:
Alpaca\footnote{\url{https://huggingface.co/datasets/yahma/alpaca-cleaned}}~\cite{taori2023alpaca}, Dolly-15K\footnote{\url{https://huggingface.co/datasets/databricks/databricks-dolly-15k}}~\cite{conover2023free}, and NI-20~\cite{wang-etal-2022-super, kadasi2025adaptlearningtaskmixtures} (a subset of Natural Instructions with 20 task types), refer Table~\ref{tab:dataset} for more details.
Each training instance is a triple $(I,X,Y)$ consisting of an instruction, an input, and a target response.

\subsection{Models}
We evaluate across three open-weight LLM families:
Gemma-3-1B-PT\footnote{\url{https://huggingface.co/google/gemma-3-1b-pt}}~\cite{gemmateam2025gemma3technicalreport}, LLaMA-3.2-1B\footnote{\url{https://huggingface.co/meta-llama/Llama-3.2-1B}}~\cite{grattafiori2024llama3herdmodels, meta2024llama32}, and Qwen3-0.6B\footnote{\url{https://huggingface.co/Qwen/Qwen3-0.6B-Base}}~\cite{yang2025qwen3technicalreport}.
We use the corresponding 12B instruct variants of these model families to generate alternative instructions, while using the 1B base models for $\tss$/$\tsspp$ scoring.
We fine-tune each 1B model under controlled budget settings.

\subsection{Budget Settings}
We evaluate training budgets as dataset fractions
$\rho \in \{0.05, 0.15, 0.45, 0.75, 1.0\}$.
For each dataset--model pair and each $\rho$, we train using:
(i) a \textbf{filtered subset} of size $\rho|\mathcal{D}|$ (for subset-selection methods), or
(ii) the \textbf{full dataset with per-example weights} (for weighting methods), matching the setting used by that variant.

\subsection{Data Selection and Weighting Variants}
We compare $\tss$ and $\tsspp$ variants with the following data construction strategies (definitions are provided in Section~\ref{sec:method}):
\begin{itemize}[nolistsep]
    \item \textbf{Random}: uniform random subset at each budget $\rho$.
    \item \textbf{PPL}: perplexity-based ranking for subset selection~\cite{li-etal-2024-superfiltering}.
    \item \textbf{IFD}: instruction-following difficulty ranking~\cite{li-etal-2024-quantity}.
\end{itemize}

\paragraph{Implementation note.}
All selection/weighting scores are computed using a \emph{frozen} scoring model, and we keep the scoring procedure fixed across budgets
for a given dataset--model setting.
For compute reasons, we set the number of generated alternatives to $K=4$ per instance when constructing $\tss$/$\tsspp$ scores.
Training then uses the resulting subset or weights.

\subsection{Evaluation and Experimental Setting}
We study budgeted instruction tuning across different dataset fractions $\rho \in \{0.05, 0.15, 0.45, 0.75, 1.0\}$ across three training datasets (Alpaca, Dolly-15K, and NI-20 and three decoder-only models (Gemma, Llama, and Qwen). All trained checkpoints are evaluated using the LM-Eval Harness\footnote{\url{https://github.com/EleutherAI/lm-evaluation-harness}}~\cite{biderman2024lessonstrenchesreproducibleevaluation} on four downstream benchmarks: \textsc{ARC-C}~\cite{clark2018thinksolvedquestionanswering}, \textsc{MMLU}~\cite{hendrycks2021measuring}, \textsc{HellaSwag}~\cite{zellers-etal-2019-hellaswag}), and \textsc{TruthfulQA}~\cite{lin-etal-2022-truthfulqa}. We report individual benchmark scores as well as their (\textbf{SUM}), which serves as our primary aggregate metric (higher is better).

We further compare $\tss$ and $\tsspp$ and their weighted exponent variants\footnote{In the rest of the paper, we denote it as $\tss$ (E) and $\tsspp$ (E).} with Random fraction baseline, quality baselines such as PPL and IFD.

\section{Results}
\label{sec:results}

\begin{table*}[t]
\centering
\small
\setlength{\tabcolsep}{5pt}
\begin{tabular}{llrrrrl}
\toprule
Model & Dataset & Random & Best & $\Delta$ & Best method \\
\midrule
Gemma & Alpaca & 132.758 & \textbf{132.902} & +0.144 & $\tss$ \\
Gemma & Dolly & 130.091 & \textbf{130.950} & +0.859 & $\tsspp$ (E) \\
Gemma & NI-20 & 130.491 & \textbf{132.137} & +1.646 & $\tss$ \\
\midrule
Llama & Alpaca & 141.921 & \textbf{142.704} & +0.783 & $\tsspp$ (E) \\
Llama & Dolly & 140.270 & \textbf{140.573} & +0.303 & $\tsspp$ \\
Llama & NI-20 & 140.642 & \textbf{140.642} & +0.000 & Random \\
\midrule
Qwen & Alpaca & 156.415 & \textbf{157.162} & +0.747 & $\tsspp$ (E) \\
Qwen & Dolly & 152.514 & \textbf{153.582} & +1.068 & $\tsspp$ (E) \\
Qwen & NI-20 & 154.376 & \textbf{155.153} & +0.777 & $\tsspp$ (E) \\
\bottomrule
\end{tabular}
\caption{Budgeted subset selection at $5\%$ retention. ``Random'' uses a random subset of equal size; ``Best'' is the highest SUM among all selection/weighting strategies at this budget. $\Delta$ is computed w.r.t. Random. SUM is the unweighted sum of ARC-C, MMLU, HellaSwag, and TruthfulQA.}
\label{tab:budget5}
\end{table*}

\subsection{Small-Budget Regime: $5\%$ and $15\%$ Retention}
Tables~\ref{tab:budget5} and~\ref{tab:budget15} summarize, for each (model, dataset), the best-performing strategy at $\rho=0.05$ and $\rho=0.15$, reported as \textbf{SUM} along with $\Delta$ vs.\ Random at the same budget.

\paragraph{$5\%$ retention\footnote{We use the term \emph{retention fraction} to emphasize that each method deterministically selects and trains on a fixed subset of the original supervision, rather than resampling data dynamically during training.}.}
At $\rho=0.05$, the best strategy improves over Random in \textbf{8/9} settings, with an average gain of \textbf{+0.703 SUM}. The largest gain is \textbf{+1.646 SUM} for Gemma on \textsc{NI-20} (\textbf{132.137} vs.\ 130.491; Table~\ref{tab:budget5}). 
Across datasets, \tsspp{} is the most frequent winner at this budget (\textbf{6/9} settings), indicating that hard-negative contrast plus a quality term is particularly robust when only a tiny subset of training data can be retained. Vanilla \tss{} wins \textbf{2/9}, while Random wins \textbf{1/9}.

\begin{table*}[t]
\centering
\small
\setlength{\tabcolsep}{5pt}
\begin{tabular}{llrrrrl}
\toprule
Model & Dataset & Random & Best & $\Delta$ & Best method \\
\midrule
Gemma & Alpaca & 134.250 & \textbf{134.250} & +0.000 & Random \\
Gemma & Dolly & 130.724 & \textbf{132.231} & +1.507 & $\tss$ (E) \\
Gemma & NI-20 & 129.941 & \textbf{131.834} & +1.893 & $\tss$ \\
\midrule
Llama & Alpaca & 141.825 & \textbf{144.603} & +2.778 & $\tsspp$ \\
Llama & Dolly & 140.234 & \textbf{140.646} & +0.412 & PPL \\
Llama & NI-20 & 140.717 & \textbf{140.717} & +0.000 & Random \\
\midrule
Qwen & Alpaca & 158.580 & \textbf{159.215} & +0.635 & IFD \\
Qwen & Dolly & 152.303 & \textbf{154.110} & +1.807 & $\tsspp$ (E) \\
Qwen & NI-20 & 151.884 & \textbf{154.765} & +2.881 & $\tss$ (E) \\
\bottomrule
\end{tabular}
\caption{Budgeted subset selection at $15\%$ retention (same conventions as Table~\ref{tab:budget5}).}
\label{tab:budget15}
\end{table*}

\paragraph{$15\%$ retention.}
At $\rho=0.15$, the best strategy improves in \textbf{6/9} settings with an average gain of \textbf{+1.253 SUM}. The maximum improvement is \textbf{+2.881 SUM} for Qwen on \textsc{NI-20} (\textbf{154.765} vs.\ 151.884), achieved by \tss{} with exponential weighting (Table~\ref{tab:budget15}). In this regime, the winner set becomes more diverse: Random remains competitive (\textbf{3/9} wins), \tss{} wins \textbf{3/9}, \tsspp{} wins \textbf{2/9}, and a quality heuristic wins \textbf{1/9} (PPL for Llama on Dolly; Table~\ref{tab:budget15}). This pattern suggests that as the budget relaxes, quality signals can catch up, but instruction-specificity remains a strong and often complementary axis of selection.

\subsection{Budget Sweeps: Crossovers and ``Sweet spots''}
\begin{figure}[htbp]
  \centering

  \begin{subfigure}{0.98\linewidth}
    \centering
    \includegraphics[width=\linewidth]{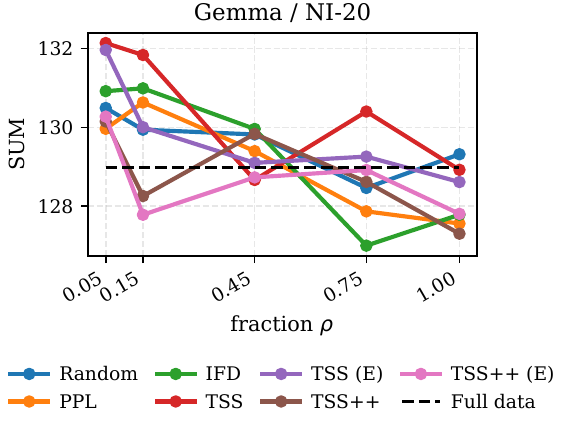}
    \caption{}\label{fig:budget_sweep_a}
  \end{subfigure}\par\vspace{0.6em}

  \begin{subfigure}{0.98\linewidth}
    \centering
    \includegraphics[width=\linewidth]{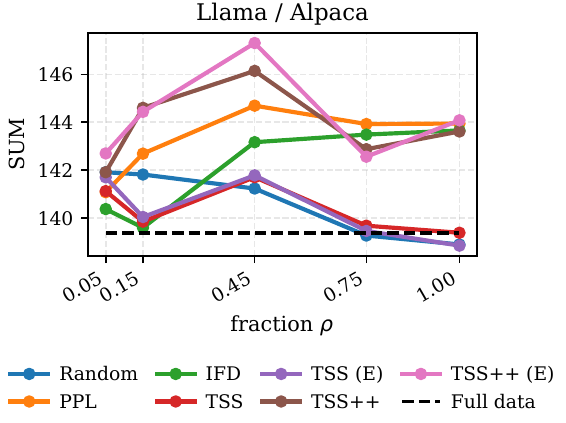}
    \caption{}\label{fig:budget_sweep_b}
  \end{subfigure}\par\vspace{0.6em}

  \begin{subfigure}{0.98\linewidth}
    \centering
    \includegraphics[width=\linewidth]{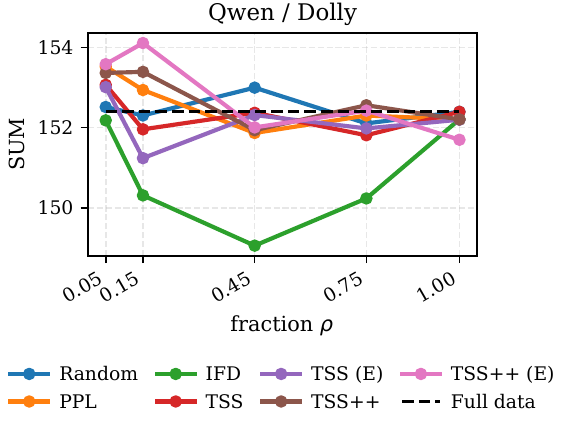}
    \caption{}\label{fig:budget_sweep_c}
  \end{subfigure}

  \caption{\textbf{Budget sweeps.} X-axis: retention fraction $\rho \in \{0.05,0.15,0.45,0.75,1.0\}$. Y-axis: SUM. Plot one curve per method (Random, PPL, IFD, $\tss$, $\tss$ (E), $\tsspp$, $\tsspp$ (E)).}
  \label{fig:budget_sweep}
\end{figure}

The full sweeps (Figure~\ref{fig:budget_sweep}) reveal two consistent patterns across models and datasets, and the three representative plots included in the paper (Gemma--\textsc{NI-20}, Llama--Alpaca, Qwen--Dolly) illustrate them clearly: in each case, an instruction-specific method (\tss{} or \tsspp{}) forms the top-performing curve over the sweep and attains the best operating point among the compared baselines.


\begin{table*}[t]
\centering
\small
\setlength{\tabcolsep}{4pt}
\begin{tabular}{llrrrrl r}
\toprule
Model & Dataset & Zero-shot & Full SFT & Best overall & $\Delta$ vs Full & Method (budget) \\
\midrule
Gemma & Alpaca & 129.923 & 135.176 & \textbf{136.089} & +0.913 & Random (0.75) \\
Gemma & Dolly & 129.923 & 131.509 & \textbf{132.526} & +1.017 & Random (0.75) \\
Gemma & NI-20 & 129.923 & 128.977 & \textbf{132.137} & +3.160 & $\tss$ (0.05) \\
\midrule
Llama & Alpaca & 127.493 & 139.388 & \textbf{147.304} & +7.916 & $\tsspp$ (E) (0.45) \\
Llama & Dolly & 127.493 & 138.191 & \textbf{141.979} & +3.788 & PPL (1.0) \\
Llama & NI-20 & 127.493 & 132.912 & \textbf{140.717} & +7.805 & Random (0.15) \\
\midrule
Qwen & Alpaca & 154.218 & 156.843 & \textbf{159.215} & +2.372 & IFD (0.45) \\
Qwen & Dolly & 154.218 & 152.394 & \textbf{154.110} & +1.716 & $\tsspp$ (E) (0.15) \\
Qwen & NI-20 & 154.218 & 151.595 & \textbf{155.153} & +3.558 & $\tsspp$ (E) (0.05) \\
\bottomrule
\end{tabular}
\caption{Best configuration across all budgets and selection strategies (SUM). ``Full SFT'' corresponds to training on the entire dataset; ``Zero-shot'' is evaluated without instruction tuning. ``Method (budget)'' reports the winning selection heuristic and the retained fraction.}
\label{tab:overall}
\end{table*}

\paragraph{(i) Intermediate retention can outperform full-dataset tuning.}
Table~\ref{tab:overall} reports the best configuration across \emph{all} budgets and selection strategies.
In \textbf{8/9} cases, the best overall result occurs at $\rho<1.0$, i.e., discarding some training instances improves generalization.
For example, Llama on Alpaca peaks at \textbf{147.304} SUM with \tsspp{} (E) at $\rho=0.45$, outperforming full-dataset tuning by \textbf{+7.916 SUM}.
Similarly, Qwen on Alpaca peaks at \textbf{159.215} SUM (IFD, $\rho=0.45$), a \textbf{+2.372 SUM} gain over full-dataset tuning.
These results indicate that even curated instruction datasets contain a substantial fraction of redundant or low-utility instances, and that selection can beat na\"{i}ve ``use everything'' scaling.

\paragraph{(ii) The best heuristic is model- and dataset-dependent, but instruction-specificity wins often.}
While \tsspp{} dominates the extreme small-budget regime (Table~\ref{tab:budget5}), quality heuristics become competitive at higher retention for certain (model, dataset) pairs (e.g., PPL winning Llama--Dolly at $\rho=0.15$; Table~\ref{tab:budget15}, and IFD winning Qwen--Alpaca overall; Table~\ref{tab:overall}). At the same time, the representative sweeps in Figure~\ref{fig:budget_sweep} show that \tss{}/\tsspp{} can remain the top curve across substantial portions of the budget range, motivating reporting full sweeps rather than a single operating point.

\subsection{Winner heatmap across budgets}

\begin{figure}[htbp]
\centering
\includegraphics[width=\linewidth]{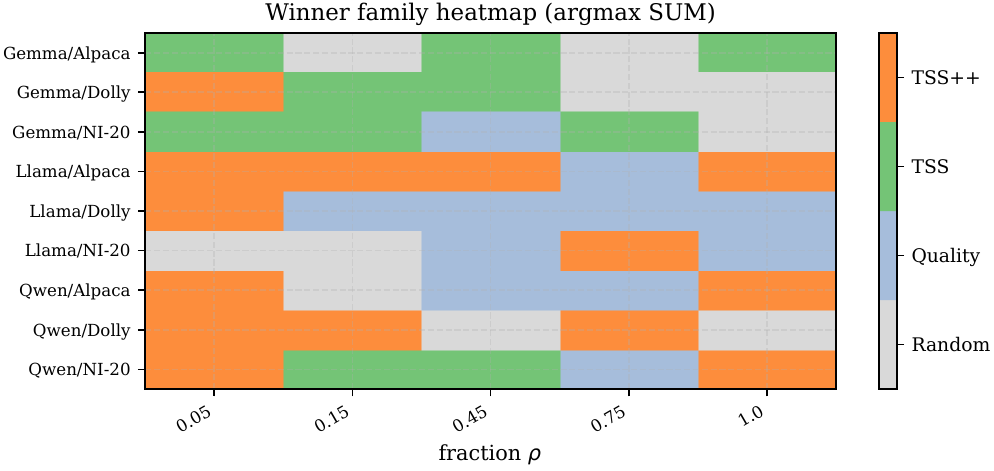}
\caption{\textbf{Winner map across budgets.} A heatmap with rows as (model, dataset) and columns as $\rho$; each cell indicates the winning method family (Random vs Quality vs $\tss$ vs $\tsspp$).}
\label{fig:winner_map}
\end{figure}

Figure~\ref{fig:winner_map} provides a compact ``winner map'' across all budgets: rows correspond to (model, dataset) pairs and columns to $\rho$, with each cell marking the winning \emph{method family} (Random vs.\ Quality vs.\ \tss{} vs.\ \tsspp{}).
Two takeaways stand out.
First, the heatmap makes the \emph{budget dependence} explicit: instruction-specificity methods occupy a large portion of the low-to-mid $\rho$ region, while Random/quality methods appear more frequently at higher $\rho$ for certain rows.
Second, the map highlights that crossovers are not random noise: the same (model, dataset) can shift families as $\rho$ increases, consistent with the intuition that specificity is most valuable when the budget is tight, whereas quality and coverage become increasingly important as more data is retained.

\subsection{Where do the gains come from? Benchmark-level breakdown}
To check whether improvements in SUM are driven by a single benchmark or reflect broader generalization, we analyze a representative high-gain setting (Llama, Alpaca, $\rho=0.45$).
Table~\ref{tab:rep_llama_alpaca} reports per-benchmark scores.

\begin{table}[tb]
\centering
\small
\setlength{\tabcolsep}{5pt}
\begin{tabular}{lrrr}
\toprule
Benchmark & Random & Best & $\Delta$ \\
\midrule
ARC-C & 32.594 & \textbf{33.959} & +1.365 \\
MMLU & 34.831 & \textbf{37.423} & +2.592 \\
HellaSwag & 48.357 & \textbf{48.994} & +0.637 \\
TruthfulQA & 25.459 & \textbf{26.928} & +1.469 \\
SUM & 141.241 & \textbf{147.304} & +6.063 \\
\bottomrule
\end{tabular}
\caption{Per-benchmark breakdown for a representative high-gain setting: Llama on Alpaca at $45\%$ retention. ``Best'' corresponds to $\tsspp$ (E). All four benchmarks improve simultaneously, indicating that gains are not driven by a single dataset.}
\label{tab:rep_llama_alpaca}
\end{table}

The winning \tsspp{} configuration improves all four benchmarks simultaneously (ARC-C, MMLU, HellaSwag, TruthfulQA), indicating that the gain is not benchmark-specific, suggesting that selection is not trading one benchmark for another but improving generalization more broadly.

\paragraph{A note on weighting vs.\ filtering.}
Across the small-budget tables, exponential weighting variants appear among the winners in several settings (e.g., multiple $\tsspp$ (E) wins at $\rho=0.05$; Table~\ref{tab:budget5}), while at $\rho=0.15$ the strongest gain is achieved by \tss{} with exponential weighting on Qwen--\textsc{NI-20} (Table~\ref{tab:budget15}).
This suggests that when the retained subset is very small or when the dataset mixes strong and weak supervision signals, smoothly emphasizing high-specificity examples can be competitive with hard top-$\rho$ filtering, although the best choice remains model/dataset dependent.

\section{Limitations}
\label{sec:limitations}
Our study has several limitations. 
First, \textsc{$\tss$} and \textsc{$\tsspp$} are \emph{model-dependent}: scores are computed using a frozen LM, so the ranking can change with the choice of scoring model, its calibration, and its prompt/template. While we evaluate across three decoder-only models and observe consistent trends, we do not claim model-invariant specificity estimates.
Second, \textsc{$\tss$} relies on generating or retrieving plausible alternative instructions. If the alternative set is low-quality (e.g., repetitive, implausible, or overly generic), \textsc{$\tss$} may overestimate specificity; conversely, if alternatives are too strong or off-distribution, it may underestimate specificity. Although \textsc{$\tsspp$} mitigates ``easy-negative'' effects via hard negatives, the method remains sensitive to the alternative construction pipeline (generation prompts, retrieval corpus, embedding model, and hyperparameters).
Third, scoring is computationally non-trivial: computing per-example likelihoods for $K$ alternatives (and hard-negative mining) can be expensive for large corpora, even though it requires only forward passes. This cost may limit applicability in very large-scale instruction pools without additional approximation (e.g., caching, smaller scoring models, or reduced $K$).
Fourth, our experiments focus on three instruction pools (Alpaca, Dolly-15K, NI-20, three small open models, and four downstream benchmarks. The conclusions may not fully transfer to larger models, other domains (e.g., multimodal or tool-use data), or alternative evaluation suites. In particular, \textbf{SUM} is an unweighted aggregate over a small benchmark set; different benchmark mixes or weighting schemes may change the observed trade-offs.
Finally, \textsc{$\tss$} targets instruction-specificity and does not directly guarantee factuality, harmlessness, or alignment of the resulting tuned models. Data selection may improve generalization under a budget but does not replace safety-specific training or evaluation.

\section{Ethical Considerations}
\label{sec:ethics}
This work proposes a data scoring and selection framework for instruction tuning.
Because our approach can \emph{amplify} certain subsets of data, it may also amplify biases present in the underlying instruction pools or the scoring model. For example, if the frozen LM assigns systematically higher likelihood to particular dialects, writing styles, or demographic topics, \textsc{$\tss$}-based selection could disproportionately retain such examples. We recommend applying standard dataset governance practices (documentation, bias analysis, and auditing) and monitoring distributional shifts introduced by selection.
In addition, our method leverages existing public instruction datasets that may contain problematic content (toxicity, stereotypes, sensitive information, or licensing restrictions). We do not introduce new personal data collection, but we encourage users of \textsc{$\tss$}/\textsc{$\tsspp$} to respect dataset licenses and to apply content filtering where appropriate.
From a misuse perspective, improved data efficiency can lower the barrier to tuning models for harmful purposes. While \textsc{$\tss$} is intended for research on data quality and instruction dependence, we advise that deployments incorporate safety mitigation (e.g., refusal training, red teaming, and usage policies) and that release artifacts avoid enabling targeted harmful fine-tuning.
Finally, there are environmental and resource considerations: although \textsc{$\tss$} uses only forward passes, large-scale scoring can still incur substantial compute and energy consumption. We report compute details and encourage practitioners to reuse cached scores, prefer smaller scoring models when possible, and reduce $K$ or hard-negative mining cost under limited resources.

\section{Conclusion}
\label{sec:conclusion}
We introduced \textbf{Task--Specificity Score ($\tss$)}, a simple, model-based measure of whether a target output $Y$ is \emph{uniquely compatible} with the given instruction $I$ among plausible alternative instructions for the same input $X$. \textsc{$\tss$} operationalizes instruction dependence as a contrast between the likelihood under the true instruction and an empirical estimate of $p(Y\mid X)$ formed by averaging over alternatives. We further proposed \textbf{\textsc{$\tsspp$}}, which strengthens specificity estimation by focusing on hard alternatives via an InfoNCE-style contrastive objective and by adding an explicit quality term to remain competitive with strong quality-oriented filters.
Across three instruction pools and three open LLMs, we find that prioritizing high-specificity examples improves downstream performance under tight retention budgets and complements quality-based selection methods, highlighting that \emph{instruction sensitivity} is a distinct and practically useful signal beyond generic likelihood.
More broadly, our results suggest that data curation for instruction tuning should consider not only whether outputs are ``easy'' or ``likely,'' but also whether the instruction meaningfully constrains the supervision signal.


\bibliography{tacl2021, anthology-1, anthology-2, custom}
\bibliographystyle{acl_natbib}

\onecolumn
\appendix
\section{Prompts}
\label{sec:prompts}

\begin{tcolorbox}[
  title={Alternative Instructions Generator Prompt},
  colback=white,
  colframe=black,
  coltitle=white,
  colbacktitle=black,
  fonttitle=\bfseries,
  boxrule=0.6pt,
  arc=2pt,
  left=6pt,right=6pt,top=6pt,bottom=6pt,
]

\textbf{SYSTEM (when input text is available).}
\vspace{0.25em}
\begin{lstlisting}[style=promptstyle]
Generate exactly {k} distinct task instructions for the input below.
Each instruction must:
- Be answerable using only that input (no external documents).
- Ask for a different type of output than the others.
- Be a single sentence, at most 25 words.

Return a numbered list (1., 2., ...) with one instruction per line.
Do not include any explanations, headers, or extra text.
\end{lstlisting}

\textbf{USER}
\vspace{0.25em}
\begin{lstlisting}[style=promptstyle]
Input:
{input_text}
\end{lstlisting}

\tcbline

\textbf{SYSTEM (when input text is empty).}
\vspace{0.25em}
\begin{lstlisting}[style=promptstyle]
Given the original task instruction below, generate exactly {k} distinct alternative
task instructions that do NOT require any extra input text.

Each alternative instruction must:
- Be about the same general topic or domain as the original instruction.
- Ask for a different type of output or perspective (not just a paraphrase).
- Be a single sentence, at most 25 words.

Return a numbered list (1., 2., ...) with one instruction per line.
Do not include any explanations, headers, or extra text.
\end{lstlisting}

\textbf{USER}
\vspace{0.25em}
\begin{lstlisting}[style=promptstyle]
Original instruction:
{original_instruction}
\end{lstlisting}

\end{tcolorbox}

\end{document}